\documentclass[sigconf]{acmart}
\usepackage{enumitem}
\setlist{nosep}

\AtBeginDocument{%
  }
    
\copyrightyear{2025}
\acmYear{2025}
\setcopyright{acmlicensed}\acmConference[MM '25]{Proceedings of the 33rd ACM International Conference on Multimedia}{October 27--31, 2025}{Dublin, Ireland}
\acmBooktitle{Proceedings of the 33rd ACM International Conference on Multimedia (MM '25), October 27--31, 2025, Dublin, Ireland}
\acmDOI{10.1145/3746027.3762093}
\acmISBN{979-8-4007-2035-2/2025/10}

\begin{document}

\makeatletter
\def\@ACM@checkaffil{
    \if@ACM@instpresent\else
    \ClassWarningNoLine{\@classname}{No institution present for an affiliation}%
    \fi
    \if@ACM@citypresent\else
    \ClassWarningNoLine{\@classname}{No city present for an affiliation}%
    \fi
    \if@ACM@countrypresent\else
        \ClassWarningNoLine{\@classname}{No country present for an affiliation}%
    \fi
}
\makeatother

\title{Multi-Level CLS Token Fusion for Contrastive Learning in Endoscopy Image Classification}

\author{Y Hop Nguyen}
\orcid{0009-0001-0021-5429}
\affiliation{%
  \institution{University of Science, VNU-HCM}
  \city{Ho Chi Minh}
  \country{Vietnam}
}
\affiliation{%
  \institution{South Telecom JSC}
  \city{Ho Chi Minh}
  \country{Vietnam}
}
\email{hop.nguyen@southtelecom.vn}

\author{Doan Anh Phan Huu}
\orcid{0009-0008-4512-0053}
\affiliation{%
  \institution{South Telecom JSC}
  \city{Ho Chi Minh}
  \country{Vietnam}
}
\email{anh.phan@southtelecom.vn}

\author{Trung Thai Tran}
\email{thai.tran@southtelecom.vn}
\affiliation{%
  \institution{South Telecom JSC}
  \city{Ho Chi Minh}
  \country{Vietnam}
}
\orcid{0009-0002-1422-9685}

\author{Nhat Nam Mai}
\orcid{0009-0005-3310-7562}
\affiliation{%
  \institution{South Telecom JSC}
  \city{Ho Chi Minh}
  \country{Vietnam}
}
\email{nam.mai@southtelecom.vn}

\author{Van Toi Giap}
\orcid{0009-0007-0219-1735}
\affiliation{%
  \institution{South Telecom JSC}
  \city{Ho Chi Minh}
  \country{Vietnam}
}
\email{toi.giap@southtelecom.vn}

\author{Thao Thi Phuong Dao}
\orcid{0000-0002-0109-1114}
\affiliation{%
  \institution{University of Science, VNU-HCM}
  \city{Ho Chi Minh}
  \country{Vietnam}
}
\affiliation{%
  \institution{Thong Nhat Hospital}
  \city{Ho Chi Minh}
  \country{Vietnam}
}
\email{dtpthao@selab.hcmus.edu.vn}

\author{Trung-Nghia Le}
\orcid{0000-0002-7363-2610}
\affiliation{%
  \institution{University of Science, VNU-HCM}
  \city{Ho Chi Minh}
  \country{Vietnam}
}
\email{ltnghia@fit.hcmus.edu.vn}
\authornote{Corresponding author}

\renewcommand{\shortauthors}{Y Hop Nguyen et al.}

\begin{abstract}
We present a unified vision-language framework tailored for ENT endoscopy image analysis that simultaneously tackles three clinically-relevant tasks: image classification, image-to-image retrieval, and text-to-image retrieval. Unlike conventional CNN-based pipelines that struggle to capture cross-modal semantics, our approach leverages the CLIP ViT-B/16 backbone and enhances it through Low-Rank Adaptation, multi-level CLS token aggregation, and spherical feature interpolation. These components collectively enable efficient fine-tuning on limited medical data while improving representation diversity and semantic alignment across modalities. To bridge the gap between visual inputs and textual diagnostic context, we introduce class-specific natural language prompts that guide the image encoder through a joint training objective combining supervised classification with contrastive learning. We validated our framework through participation in the ACM MM’25 ENTRep Grand Challenge, achieving 95\% accuracy and F1-score in classification, Recall@1 of 0.93 and 0.92 for image-to-image and text-to-image retrieval respectively, and MRR scores of 0.97 and 0.96. Ablation studies demonstrated the incremental benefits of each architectural component, validating the effectiveness of our design for robust multimodal medical understanding in low-resource clinical settings.

\end{abstract}



\maketitle

\section{Introduction}

Ear, nose, and throat (ENT) endoscopy is a vital imaging modality in otolaryngology that enables direct visualization of critical anatomical structures such as the nasal cavity, pharynx, vocal cords, and auditory canal. These procedures support early diagnosis of a wide range of conditions including infections, tumors, and structural abnormalities. However, the interpretation of endoscopic images remains a highly specialized task, demanding significant clinical expertise and often suffering from inter-observer variability and diagnostic subjectivity. Manual analysis is time-consuming and may lead to inconsistent conclusions across practitioners, particularly in resource-limited settings.

With the emergence of artificial intelligence (AI), especially in the field of computer vision, there has been a growing interest in automating diagnostic procedures to support clinical decision-making. Convolutional neural networks (CNNs), such as Inception-ResNet-v2~\cite{baldassarre2017deep}, ResNet-50~\cite{he2016deep}, and MobileNetV2~\cite{sandler2018mobilenetv2}, have been successfully applied to various medical imaging tasks including endoscopic image classification~\cite{Mukhtorov2023EndoscopicIC, Subedi2024ClassificationOE}. These models can achieve high accuracy but are typically limited to unimodal input (i.e., visual-only) and are often treated as black boxes with limited explainability. While interpretability tools like Grad-CAM~\cite{selvaraju2016grad} and Score-CAM~\cite{wang2020score} provide some level of visual explanation, they still fall short in offering semantic, textual justifications or enabling natural language interactions.

More critically, these CNN-based approaches neglect the rich diagnostic information that often accompanies clinical images in textual form, such as anatomical labels, descriptions of pathological findings, or physician notes. This limits their utility in scenarios requiring cross-modal reasoning, such as querying medical databases via text, generating textual reports from images, or grounding visual observations in expert-level semantics. These limitations underscore the need for models that can jointly reason over both visual and textual modalities.

To address these challenges, we propose a unified vision-language framework for ENT endoscopy analysis that leverages the power of multimodal learning. 
Our method enhances the CLIP ViT-B/16 backbone through a combination of class-specific semantic prompts, which inject domain knowledge into the encoder; LoRA-based contrastive adaptation, which enables efficient fine-tuning under limited data; multi-level CLS token aggregation, which captures both fine-grained textures and global semantic cues; and spherical feature augmentation (SFA), which diversifies embeddings while preserving semantic consistency. A joint training objective balances supervised classification with contrastive learning, ensuring both accurate prediction and robust cross-modal alignment. 
Unlike traditional CNNs, our model is capable of not only classifying ENT images but also performing retrieval tasks across modalities, such as finding similar images from a textual query or retrieving relevant descriptions given an input image.

We evaluated our model on the ACM Multimedia 2025 ENTRep Challenge dataset \cite{nguyen2025acmmultimediagrandchallenge}, a clinically curated benchmark of annotated ENT endoscopy data. Experimental results demonstrated that our approach significantly outperforms existing baselines across all three tasks: image classification, image-to-image retrieval, and text-to-image retrieval. Our findings suggested that vision-language models hold strong potential for advancing multimodal understanding in medical imaging and point toward a scalable path for AI-assisted ENT diagnostics.

The key contributions of this work are as follows:
\begin{itemize}
    \item We introduce a novel unified multimodal framework for ENT endoscopy that supports classification, image-to-image retrieval, and text-to-image retrieval in a single architecture.
    \item We participated in the ACM MM’25 ENTRep Challenge, where our method achieved state-of-the-art performance across all three tracks, demonstrating both accuracy and robustness in real-world clinical benchmarks.
\end{itemize}

\section{Related Works}




\subsection{Feature Fusion in Vision Transformers}

Feature fusion across transformer layers has been explored to enrich hierarchical representation. While the original ViT \cite{dosovitskiy2020image} outputs features from the final transformer block, many tasks benefit from multi-level aggregation. Token-based fusion strategies such as TokenFusion \cite{rao2022tokenfusion} or TransFG \cite{he2021transfg} demonstrate that combining [CLS] tokens from intermediate layers improves performance in fine-grained recognition tasks. This motivates our design of multi-level CLS token aggregation, which captures both low-level textures and high-level semantic cues, crucial in medical image interpretation.

\subsection{Contrastive Learning in Medical Imaging}

Contrastive learning has emerged as a powerful self-supervised and weakly-supervised strategy in medical image representation learning. Works such as \textsc{ConVIRT} \cite{zhang2022contrastive} and \textsc{GLoRIA} \cite{huang2021gloria} leverage image-report pairs to learn cross-modal embeddings, demonstrating improvements in classification and retrieval across multiple imaging modalities. Recent methods further extend this paradigm by incorporating label-aware contrastive objectives \cite{chaitanya2020contrastive}, enabling fine-grained class separation in small datasets. Our work follows this line by using supervised contrastive loss aligned with domain-specific prompts and low-rank adaptation, tailored for ENT image-text alignment.

\section{Method}

\subsection{Overview}

\begin{figure*}[t]
  \centering
  \includegraphics[width=0.75\textwidth]{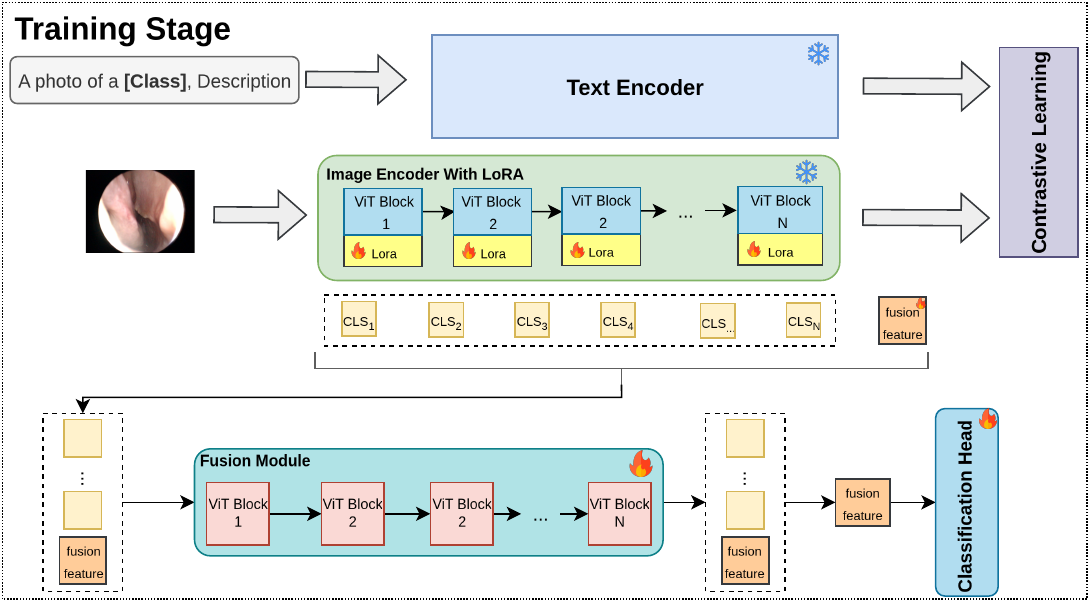}
  \caption{The overall architecture of our proposed CLIP-based multi-modal framework for ENT endoscopy analysis. The image and textual inputs are processed by a LoRA-adapted Vision Transformer (ViT) and a frozen text encoder respectively. Multi-layer CLS tokens are extracted and aggregated via a fusion module to enhance image representation. The system is trained with a hybrid objective: contrastive learning aligns vision-text modalities, while used for the classification task.}
  \label{fig:pipeline}
\end{figure*}

Our framework adopts a multi-task learning setup to jointly address image classification, image-to-image retrieval, and text-to-image retrieval. As illustrated in Figure~\ref{fig:pipeline}, it builds upon CLIP ViT-B/16 \cite{radford2021learning} and integrates four key modules to enhance visual representation learning:

\begin{itemize}
    \item \textit{Semantic Enhancement for Visual Descriptions}: Utilizes class-specific textual prompts to guide the image encoder, improving semantic discrimination for visually similar classes.

    \item \textit{Multi-level Feature Aggregation-MFA}: Combines features from multiple CLIP layers to capture both low- and high-level visual cues.

    \item \textit{LoRA-based Contrastive Learning}: Applies LoRA to the vision encoder (text encoder frozen) for efficient contrastive alignment under limited data.

    \item \textit{Slerp Feature Augmentation-SFA}: Introduces spherical interpolation in feature space to enrich embedding diversity and generalization during training.
\end{itemize}

\subsection{Semantic Enhancement for Visual Descriptions}

To enhance the semantic alignment between image and text modalities, based on CoOp \cite{zhou2022conditional}, we introduce a semantic prompting strategy that explicitly incorporates class-specific textual descriptions into the training process. This module leverages natural language to inject domain knowledge into the CLIP framework, especially beneficial for classes with high visual similarity.

Given a predefined set of classes:
\begin{align*}
\mathcal{C} = \{&\texttt{nose-right},\ \texttt{nose-left},\ \texttt{ear-right}, \\
                &\texttt{ear-left},\ \texttt{vc-open},\ \texttt{vc-closed},\ \texttt{throat} \},
\end{align*}
we define a set of enhanced textual prompts \( \tilde{c}_j \) for each class \( c_j \in \mathcal{C} \) in the following format:
\begin{equation}
\tilde{c}_j = \texttt{"A photo of a "} + c_j + \texttt{", Image description."}
\end{equation}

These prompts serve as the input to the CLIP text encoder, producing text embeddings:

\begin{equation}
u_j = \text{Encoder}_{\text{text}}(\tilde{c}_j), \quad \forall c_j \in \mathcal{C}.
\end{equation}

During the training process, the image encoder is guided by these semantic text embeddings through contrastive learning, enabling it to focus on class-specific discriminative features and the model benefits from better generalization and interpretability in downstream classification and retrieval tasks.

\subsection{LoRA-based Contrastive Learning}

To enable the CLIP image encoder to learn semantic representations from the CLIP text encoder, we apply Low-Rank Adaptation (LoRA)~\cite{hu2022lora} to each ViT block of the CLIP image encoder. LoRA is a weight update method trained using the product of two smaller matrices, denoted as \(\mathbf{A}\) and \(\mathbf{B}\), which leverages low intrinsic rank adaptation to align the model with the downstream task.

Given an input \(\mathbf{x}\), hidden state \(\mathbf{h}\), and weight matrix \(\mathbf{W} \in \mathbb{R}^{d_1 \times d_2}\), the weight adaptation process using LoRA is formulated as follows:

\begin{equation}
    \mathbf{h} = \mathbf{W} \mathbf{x} + \gamma \Delta \mathbf{W} \mathbf{x} = \mathbf{W} \mathbf{x} + \gamma \mathbf{B} \mathbf{A} \mathbf{x},
\end{equation}
where \(\mathbf{A} \in \mathbb{R}^{r \times d_2}\), \(\mathbf{B} \in \mathbb{R}^{d_1 \times r}\), and \(\Delta \mathbf{W} \in \mathbb{R}^{d_1 \times d_2}\) has a rank of \(r\), typically satisfying \(r \ll \{d_1, d_2\}\). The scalar \(\gamma = \alpha / r\) is a scaling factor, where \(r\) and \(\alpha\) are user-defined hyperparameters. The matrix \(\mathbf{A}\) is initialized using Kaiming initialization, while \(\mathbf{B}\) is initialized with zeros. This initialization strategy ensures that no updates are introduced to the weights before LoRA training, thereby preserving the original model state prior to adaptation.

LoRA is integrated into the ViT blocks of the CLIP image encoder, which consists of \(L\) stacked blocks. Each block contains a Multi-Head Attention (MHA) layer defined as:

\begin{equation}
\text{head}_i = \text{Softmax}\left(\frac{\mathbf{x} \mathbf{W}_{Q_i} (\mathbf{x} \mathbf{W}_{K_i})^T}{\sqrt{d}}\right) (\mathbf{x} \mathbf{W}_{V_i}),
\end{equation}
\begin{equation}
\text{MHA}(\mathbf{x}) = \text{concat}(\text{head}_1, \ldots, \text{head}_H)\mathbf{W}_O,
\end{equation}
where \(d\) is a scaling factor, and \(\mathbf{W}_{K_i}\), \(\mathbf{W}_{Q_i}\), \(\mathbf{W}_{V_i}\), and \(\mathbf{W}_O\) are the projection matrices for key, query, value, and output, respectively. The application of LoRA during CLIP image encoder training significantly reduces computational time and cost while maintaining high performance by enabling efficient parameter updates across all ViT blocks.

\subsection{Multi-level Feature Aggregation (MFA)}

To improve the discriminability of visual representations, we adopt a multi-level feature extraction strategy that aggregates the \texttt{[CLS]} tokens from multiple Transformer blocks within the CLIP vision encoder. Suppose the encoder consists of $L$ stacked blocks. The \texttt{[CLS]} token produced at the $l$-th block is denoted as:

\begin{equation}
\mathbf{f}^{(l)}_{\text{cls}} = \text{ViT}^{(l)}_{\text{cls}}(x), \quad l = 1, 2, \dots, L.    
\end{equation}

We collect a set of hidden features from $K$ selected layers (e.g., early, mid, late blocks):

\begin{equation}
\mathcal{F}_{\text{multi}} = \left[
\mathbf{CLS}_{\text{fusion}},\
\mathbf{f}^{(l_1)}_{\text{cls}},\ \mathbf{f}^{(l_2)}_{\text{cls}},\ \dots,\ \mathbf{f}^{(l_K)}_{\text{cls}} 
\right].
\end{equation}

These multi-level features, together with the $\mathbf{CLS}_{\text{fusion}}$ token, are then fed into a \textbf{Fusion Feature Module} based on a Transformer architecture, with the objective that the $\mathbf{CLS}_{\text{fusion}}$ token aggregates information from all input tokens

\begin{equation}
    \mathbf{CLS}_{\text{fusion}} = \text{FusionViT}(\mathcal{F}_{\text{multi}}),
\end{equation}
where $\text{FusionViT}(\cdot)$ denotes a lightweight Vision Transformer Model (ViT) \cite{dosovitskiy2020image} model that takes the set of \texttt{[CLS]} tokens as input and outputs a unified representation via attention-based fusion.

This strategy enables the model to capture both low-level visual cues (e.g., texture, edges) and high-level semantic information (e.g., object structures, anatomical patterns), improving performance in fine-grained classification tasks.

\subsection{Slerp Feature Augmentation (SFA)}

To enhance the diversity of extracted features, we generate new feature from pairs of normalized feature embeddings \(f_1, f_2 \in \mathbb{R}^{d}\) belonging to the same class via spherical linear interpolation (Slerp). The interpolated feature \(f_\text{aug}\) is defined as:

\begin{equation}
    f_\text{aug}(f_1, f_2; \lambda) =
\frac{\sin[(1 - \lambda) \theta]}{\sin \theta} f_1 + 
\frac{\sin(\lambda \theta)}{\sin \theta} f_2
\end{equation}
where \(f_1, f_2 \in \mathbb{R}^d\) are unit vectors: \(\|f_1\| = \|f_2\| = 1\), \(\lambda \in [0, 1]\) is the interpolation coefficient, \(\theta = \arccos(f_1^\top f_2)\) is the angle between the two vectors.

This interpolation preserves vector norms and ensures the trajectory lies on the unit hypersphere, effectively generating semantically consistent yet diverse features. By augmenting the feature space in this manner, the model is encouraged to generalize better during training.

\subsection{Loss function}

To guide the CLIP image encoder in learning discriminative and semantically aligned features with respect to textual descriptions, we adopt a contrastive learning objective. Specifically, we first compute the text embedding \( u \) and the image embedding \( v \) as follows:

\begin{equation}\label{eqn-5}
u_j = \text{Encoder}_{\text{text}}(\tilde{c}_j), \quad v_j = \text{Encoder}^{\text{LoRA}}_{\text{img}}(x_j),
\end{equation}
where \(\text{Encoder}_{\text{text}}(\cdot)\) denotes the frozen text encoder, and \(\text{Encoder}^{\text{LoRA}}_{\text{img}}(\cdot)\) represents the image encoder fine-tuned using LoRA. Given a batch of paired image-text samples, we compute the contrastive loss as:

\begin{equation}\label{eqn-7}
\mathcal{L}_{\text{clip contrastive}} = \frac{1}{2} \left( \mathcal{L}_{v \rightarrow u} + \mathcal{L}_{u \rightarrow v} \right),
\end{equation}

\begin{equation}\label{eqn-8}
\mathcal{L}_{v \rightarrow u} = -\frac{1}{N} \sum_{i=1}^{N} \log \frac{\exp(v_i^\top u_i)}{\sum_{j=1}^{N} \exp(v_i^\top u_j)},
\end{equation}

\begin{equation}\label{eqn-9}
\mathcal{L}_{u \rightarrow v} = -\frac{1}{N} \sum_{i=1}^{N} \log \frac{\exp(u_i^\top v_i)}{\sum_{j=1}^{N} \exp(u_i^\top v_j)},
\end{equation}
where \(N\) is the batch size. This symmetric contrastive loss encourages matched image-text pairs to have high cosine similarity, while pushing apart mismatched pairs in the embedding space, thereby improving cross-modal alignment.

Finally, we apply a linear classifier on top of the fused features to perform the image classification task. The Cross-entropy loss \cite{goodfellow2016deep} is used as the primary objective for classification, denoted as \( \mathcal{L}_{\text{classification}} \). The final training loss is a weighted combination of all loss components:

\begin{align}\label{eqn-10}
\mathcal{L}_{\text{total}} =\ & \mu_{1} \mathcal{L}_{\text{classification}} 
+ \mu_{2} \mathcal{L}_{\text{contrastive}}, 
\end{align}
where \(\mu_{1}, \mu_{2} \) are hyperparameters that control the relative contribution of each loss term during training.

\subsection{Retrieval Task}

Beside the classification task, our proposed method is also applicable to the retrieval task, including both Image-to-Image Retrieval and Text-to-Image Retrieval. For \textit{Image-to-Image Retrieval}, we extract image features using the CLIP image encoder (Equation~\ref{eqn-5}). Then, we compute the cosine similarity between the features to retrieve the most similar image from the database. For \textit{Text-to-Image Retrieval}, we extract feature representation of the text query using the CLIP text encoder (Equation~\ref{eqn-5}). The similarity between the text and the images is then computed using the formula:
\[
\text{sim}(u_i, v_j) = \text{softmax}(u_i v_j^\top),
\]
where $u_i$ is the feature vector of the text and $v_j$ is the feature vector of the images.

\section{Experiments}

\subsection{Settings}

\textbf{Dataset:} Each endoscopic image in the ENTRep Challenge dataset \cite{nguyen2025acmmultimediagrandchallenge} is annotated by ENT specialists with anatomical labels (\texttt{Classification}), condition types (\texttt{Type}), and bilingual pathological descriptions in Vietnamese (\texttt{\small Description}) and English (\texttt{\small DescriptionEN}). This ensures high clinical relevance and supports effective supervised learning. The dataset includes 565 images. Before training, black borders are removed after resizing each image to $224 \times 224$ pixels.

\vspace{0.5mm}
\textbf{Data Augmentation: } To enhance training diversity, we applied augmentations including motion blur, Gaussian blur, color/contrast adjustments, and vertical flipping. For symmetric classes (\texttt{throat}, \texttt{vc-open}, \texttt{vc-closed}), horizontal flipping was also used. Additionally, we randomly masked $16 \times 16$ patches, aligned with the CLIP ViT-B/16 patch size, covering roughly 10\% of each image.

\vspace{0.5mm}
\textbf{Training Setup: } We conducted end-to-end training for a total of 50 epochs. During training, the CLIP text encoder was frozen, and LoRA was applied to the \textbf{q, k, v} layers of all ViT blocks in the CLIP image encoder. We employed the \textit{AdamW} optimizer \cite{loshchilov2017decoupled} with a learning rate of $5 \times 10^{-4}$ and $\beta = (0.9, 0.999)$. The \textit{LoRA} parameters were set as: $r = 4$, $\alpha = 8$, dropout rate $= 0.1$. For loss weights, we set $\mu_1 = 1.0$ and $\mu_2 = 0.5$.

\subsection{Evaluation Metrics}

\textbf{  Image classification: }we report {Accuracy}, {Precision}, {Recall}, and {F1-score}, which measure overall correctness, positive predictive value, sensitivity, and their harmonic mean, respectively.

\textbf{ Image-to-image retrieval} and \textbf{text-to-image retrieval: } we adopt {Recall@1}, the proportion of queries for which the correct item is ranked first, and {Mean Reciprocal Rank (MRR)}, which reflects the average inverse rank of the first relevant item.

\section{Results}


\subsection{ENTRep Track 1: Image Classification}

\begin{table}
\centering
\caption{Image classification performance comparison.}
\label{tab:classification}
\begin{tabular}{lccccc}
\toprule
\textbf{Test Split} & \textbf{Top} & \textbf{Accuracy} & \textbf{Precision} & \textbf{Recall} & \textbf{F1} \\
\midrule
Public Test  & 3 & 0.95 & 0.95 & 0.95 & 0.95 \\
Private Test & 6 & 0.91 & 0.91 & 0.91 & 0.95 \\
\bottomrule
\end{tabular}
\end{table}

We evaluated the classification performance of our model on both the public and private test sets provided by the ENTRep Challenge. As shown in Table~\ref{tab:classification}, our method achieves high and consistent results across all standard metrics. Specifically, on the public test set, the model obtains an accuracy of 0.95, with precision, recall, and F1-score all at 0.95. On the private test set, despite increased difficulty, the model maintains strong performance with an accuracy of 0.91 and an F1-score of 0.95. These results demonstrate the model’s robustness and generalization capability across different evaluation settings.

\subsection{ENTRep Track 2: Image-to-Image Retrieval}

\begin{table}
\centering
\small
\caption{Image-to-Image Retrieval performance.}
\label{tab:img2img_retrieval}
\begin{tabular}{lccc}
\toprule
\textbf{Datatest} & \textbf{Top} & \textbf{Recall@1} & \textbf{MRR} \\

\midrule
Public Test  & 3 & 0.93 & 0.97 \\
Private Test & 2 & 0.89 & 0.94 \\

\bottomrule
\end{tabular}
\end{table}

We assess the model’s retrieval ability by evaluating its performance on image-to-image retrieval tasks using Recall@1 and Mean Reciprocal Rank (MRR). As reported in Table~\ref{tab:img2img_retrieval}, the model demonstrates strong performance in both public and private test sets. On the public test set, our method achieves a Recall@1 of 0.93 and an MRR of 0.97, indicating that relevant images are consistently retrieved at the top ranks. On the more challenging private test set, the model still maintains competitive performance with a Recall@1 of 0.89 and an MRR of 0.94. These results suggest that the model is capable of learning highly discriminative image representations for retrieval tasks.

\subsection{ENTRep Track 3: Text-to-Image Retrieval}

\begin{table}
\centering
\small
\caption{Text-to-Image Retrieval performance.}
\label{tab:text2img_retrieval}
\begin{tabular}{lccc}
\toprule
\textbf{Datatest} & \textbf{Top} & \textbf{Recall@1} & \textbf{MRR} \\

\midrule
Public Test  & 5 & 0.92 & 0.96 \\
Private Test & 4 & 0.9 & 0.95 \\

\bottomrule
\end{tabular}
\end{table}

To evaluate the model’s capability in multimodal understanding, we conduct text-to-image retrieval experiments using Recall@1 and Mean Reciprocal Rank (MRR). As shown in Table~\ref{tab:text2img_retrieval}, the model achieves a Recall@1 of 0.92 and an MRR of 0.96 on the public test set, demonstrating a strong ability to retrieve relevant visual content based on textual queries. On the private test set, the model continues to perform well, with a Recall@1 of 0.90 and an MRR of 0.95. These results confirm the effectiveness of our method in aligning textual descriptions with corresponding visual information, which is essential for multimodal medical applications.

\section{Ablation study}

\subsection{Ablation Study on Component Effectiveness}

\begin{table}
\centering
\caption{Ablation study showing performance across different architectural variants.}
\label{tab:ablation}
\resizebox{\linewidth}{!}{%
\begin{tabular}{cccc|ccc}
\toprule
\textbf{Baseline} & \textbf{LoRA} & \textbf{MFA} & \textbf{SFA} & \textbf{Acc} & \textbf{Recall@1 (i2i)} & \textbf{Recall@1 (t2i)} \\
\midrule
\checkmark &  &  &  & 0.84 & 0.85 & 0.84 \\

\checkmark & \checkmark &  &  & 0.88 & 0.89 & 0.90 \\

\checkmark & \checkmark & \checkmark &  & 0.93 & 0.91 & 0.92 \\

\checkmark & \checkmark & \checkmark & \checkmark & \textbf{0.95} & \textbf{0.93} & \textbf{0.92} \\
\bottomrule
\end{tabular}}
\end{table}

To assess the individual contributions of each component in our architecture, we conduct an ablation study by progressively adding LoRA, the Fusion Module, and the SFA module to a baseline configuration. We evaluate on public datatest, which performance across variants is summarized in Table~\ref{tab:ablation}.

Starting with the baseline, which includes only the vision-language backbone, the model achieves modest performance (Accuracy: 0.75, Recall@1 for image-to-image (i2i): 0.85, and text-to-image (t2i): 0.84). Incorporating LoRA improves all metrics significantly, confirming its effectiveness in adapting pre-trained weights for domain-specific data.

Further enhancement is observed when the fusion module is added, resulting in Accuracy of 0.93 and Recall@1 of 0.91 (i2i) and 0.92 (t2i). Finally, the full model with the SFA module achieves the highest overall performance with 0.95 Accuracy, 0.93 Recall@1 (i2i), and 0.92 Recall@1 (t2i). These results demonstrate the complementary contributions of each component and validate the design of our proposed architecture.

\subsection{Backbone Comparison}

\begin{table}
\centering
\caption{Backbone comparison performance.}
\label{tab:backbone comparison}
\begin{tabular}{lccc}
\toprule
\textbf{Model} & \textbf{Acc} & \textbf{F1} & \textbf{Params (M)} \\
\midrule
MobileNetV2~\cite{sandler2018mobilenetv2} & 0.73 & 0.72 & \textbf{3.5} \\
Incep.ResNetV2~\cite{baldassarre2017deep} & 0.81 & 0.81 & 55.9 \\
ResNet-50 \cite{he2016deep} & 078 & 0.79 & 23.5 \\
BiomedCLIP \cite{zhang2023biomedclip} & 0.92 & 0.92 & 86.7 \\
CLIP-ViT-B/16 \cite{radford2021learning} & 0.84 & 0.84 & 86.7 \\
Ours & \textbf{0.95} & \textbf{0.95} & 89.3 \\
\bottomrule
\end{tabular}
\end{table}

To assess the effect of different backbone architectures, we compare several widely used CNN \cite{o2015introduction} and Transformer-based models \cite{vaswani2017attention}, as shown in Table~\ref{tab:backbone comparison}. Lightweight models like MobileNetV2 perform poorly despite low parameter counts, while deeper CNNs such as Inception-ResNetV2 and ResNet-50 offer moderate gains.

Transformer-based backbones, especially CLIP ViT-B/16, yield better performance (84\% Acc, 84\% F1) but at a higher computational cost. Our model, which builds upon CLIP ViT-B/16 with LoRA, multi-level aggregation, and feature interpolation, achieves the best results (95\% Acc, 95\% F1), demonstrating the value of both backbone choice and domain-specific enhancements.

\subsection{Feature Representation Analysis via t-SNE}

\begin{figure}
  \centering
  \includegraphics[trim= {0 0 0 0},clip, width=\linewidth]{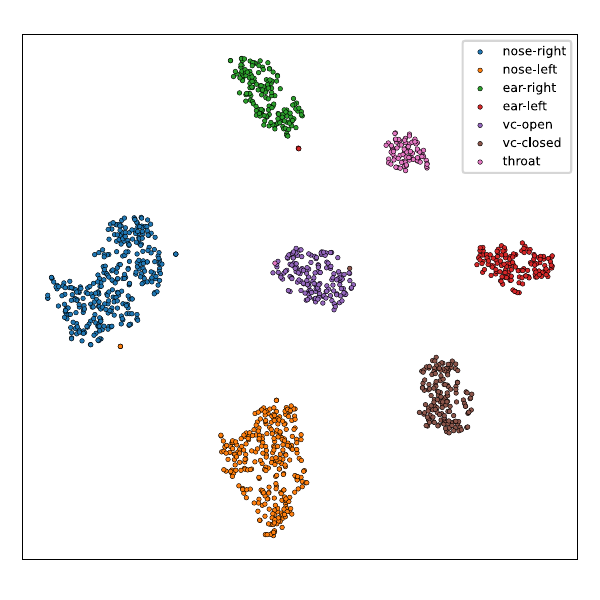}
  \caption{t-SNE visualization of the features extracted from the training set}
  \label{fig:feature}
\end{figure}

Figure \ref{fig:feature} presents the t-SNE projection of the extracted features from the training set. Each point corresponds to a sample, color-coded by its anatomical label. The clusters are generally well-separated, indicating effective representation learning. However, notable overlaps occur between the ear-left / ear-right and nose-left / nose-right pairs, suggesting confusion in features associated with symmetrical or visually similar regions. This highlights the challenge of discriminating fine-grained lateral features and suggests that these components contribute less distinctly to spatial encoding.

\subsection{Grad-CAM inspection}

\begin{figure}[t!]
  \centering
  \includegraphics[width=\linewidth]{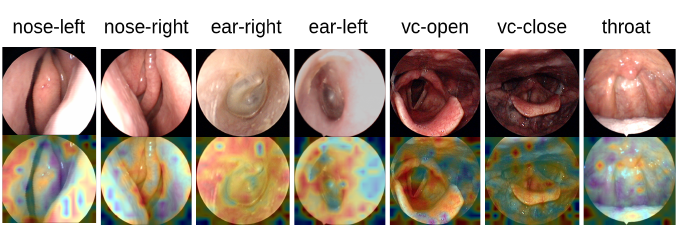}
  \caption{Grad-CAM visualization highlights the important regions learned by the model for making classification decisions.}
  \label{fig:gradcam}
\end{figure}

We performed Grad-CAM visualization in Figure \ref{fig:gradcam} and observed that the features at nose-left, nose-right, ear-left, and ear-right were learned around boundary regions. These features are sensitive to variations in the viewing angle of the observation tool, which can lead to misclassification. In contrast, the features at open-vc, close-vc, and throat were accurately learned at anatomically important positions, resulting in better classification performance.

\section{Limitations}

Despite the strong overall performance of our proposed framework, several limitations were encountered during development and experimentation. The most significant constraint lies in the dataset itself, which consists of only 565 labeled ENT endoscopic images. This limited sample size restricts the model's generalization capability, especially when handling rare or visually ambiguous cases.

A key challenge involves distinguishing anatomically similar classes, such as left and right nasal cavities (\texttt{nose-left},\texttt{nose-right}) and ear canals (\texttt{ear-left}, \texttt{ear-right}). In many cases, side-specific orientation cues are absent or inconsistent, making it difficult even for human annotators to clearly define the label. Additionally, in ear images, the endoscopic directional guides are sometimes missing, and the light source from the endoscope can cause glare or overexposure, leading to reduced visibility and potential loss of clinical detail.

Label ambiguity also emerged as a notable issue, particularly for the \texttt{vc-open} and \texttt{vc-closed} classes. These refer to the vocal cord states, but several images show partially open or not-fully-closed states, which do not clearly fit into either class. This ambiguity introduces noise into both training and evaluation. Moreover, some textual annotations in the dataset—especially those in the \texttt{DescriptionEN} field—were inconsistent or loosely defined, which reduces the reliability of supervised image-text alignment.

Lastly, endoscopic images exhibit considerable variation in angle, zoom, lighting, and insertion depth, resulting in high intra-class variability. This further complicates model learning, especially under data-limited settings and with fixed prompts. These limitations highlight the challenges of working with real-world ENT endoscopy data and the complexity of multimodal learning in a clinical context.

\section{Conclusion}
We proposed a unified vision-language framework for ENT endoscopy analysis that jointly tackles image classification, image-to-image retrieval, and text-to-image retrieval. Built on top of CLIP ViT-B/16, our method incorporates Low-Rank Adaptation (LoRA), multi-level feature aggregation, and spherical interpolation-based augmentation to enhance visual representation and cross-modal alignment in clinical settings.

Evaluated on the ENTRep Challenge benchmark, our model achieves 95\% classification accuracy, 0.93 Recall@1 for image retrieval, and 0.92 Recall@1 for text retrieval. Ablation studies confirm the complementary contributions of each architectural component, validating the overall design.

Despite strong performance, the approach is limited by the size of the dataset and the use of handcrafted textual prompts. Future directions include exploring learnable or dynamic prompting strategies, extending the model to handle temporal video data, and integrating large language models to improve diagnostic reasoning and applicability in real-world clinical environments.

\section*{Acknowledgment}

This research is funded by Vietnam National University - Ho Chi Minh City (VNU-HCM) under grant number 36-2024-44-02. 

We also acknowledge South Telecommunications \& Software Joint Stock Company (South Telecom JSC) for supporting this work.


\bibliographystyle{ACM-Reference-Format}
\balance
\bibliography{references.bib}{}

\end{document}